\documentclass[lettersize,journal]{IEEEtran}
\usepackage{amsmath,amsfonts}
\usepackage{algorithm}
\usepackage{array}
\usepackage[caption=false,font=normalsize,labelfont=sf,textfont=sf]{subfig}
\usepackage{textcomp}
\usepackage{stfloats}
\usepackage{url}
\usepackage{verbatim}
\usepackage{graphicx}
\usepackage{cite}
\usepackage{algorithmicx}
\usepackage{algpseudocode}
\usepackage{color}
\usepackage{booktabs}
\usepackage{ragged2e} 
\usepackage{makecell, multirow, tabularx}
\usepackage{utfsym}
\usepackage{amssymb}
\usepackage{subcaption}
\usepackage{MnSymbol}
\usepackage{wasysym}
\usepackage{gensymb}
\newcommand{\ZY}[1]{{\color[rgb]{0,0,0}#1}} 

\begin{document}

\title{World-Consistent Data Generation for Vision-and-Language Navigation}

\author{Yu Zhong, Zihao Zhang, Rui Zhang, Shuo Wang, Chuan Fang, Lindong Huang, Xishan Zhang \\ Jiaming Guo, Shaohui Peng, Di Huang, Yanyang Yan, Xing Hu, Qi Guo
\thanks{
Yu Zhong, Zihao Zhang, Rui Zhang, Shuo Wang, Lindong Huang, Xishan Zhang, Jiaming Guo, Di Huang, Yanyang Yan, Xing Hu, Qi Guo are with the State Key Lab (SKL) of Processors, Institute of Computing Technology (ICT), Chinese Academy of Sciences (CAS), Beijing, China, the University of Chinese Academy of Sciences, Beijing, China.

Chuan Fang is with the Hong Kong University of Science and Technology, Hong Kong.

Shaohui Peng is with the Institute of Software, CAS, Beijing, China.
}}




\maketitle

\begin{abstract}
Vision-and-Language Navigation (VLN) is a challenging task that requires an agent to navigate through photorealistic environments following natural-language instructions.
One main obstacle existing in VLN is data scarcity, leading to poor generalization performance over unseen environments.
Though data argumentation is a promising way for scaling up the dataset, how to generate VLN data both diverse and world-consistent remains problematic.
To cope with this issue, we propose the world-consistent data generation (WCGEN), an efficacious data-augmentation framework satisfying both diversity and world-consistency, aimed at enhancing the generalization of agents to novel environments.
Roughly, our framework consists of two stages, the trajectory stage which leverages a point-cloud based technique to ensure spatial coherency among viewpoints, and the viewpoint stage which adopts a novel angle synthesis method to guarantee spatial and wraparound consistency within the entire observation.
By accurately predicting viewpoint changes with 3D knowledge, our approach maintains the world-consistency during the generation procedure.
Experiments on a wide range of datasets verify the effectiveness of our method, demonstrating that our data augmentation strategy enables agents to achieve new state-of-the-art results on all navigation tasks, and is capable of enhancing the VLN agents’ generalization ability to unseen environments.
\end{abstract}


\section{Introduction}
Vision-and-Language Navigation (VLN) is a sequential decision-making task where an embodied agent is required to perform high-level navigation actions in complex environments following natural language instructions  \cite{20}. 
This research domain has recently garnered growing attention due to its potential applicability in embodied artificial intelligence and human-machine interaction.

One main challenge existing in VLN is data scarcity. 
Though numerous datasets with diverse communication complexity and objective categories have been curated, most VLN datasets are built upon the same 3D indoor environments of the RGB-D dataset Matterport3D \cite{23} that contains only 90 building-scale scenes and 10,800 panoramic images.
The reason lies in that manually collecting photorealistic observations and annotating instructions are extremely arduous and time-consuming.
Furthermore, given the inherent complexity of VLN tasks under real-world settings, there is often a pronounced mismatch between the training and inference scenarios.
Thus, data scarcity can severely hamper the generalization of VLN agents, restricting their applicability to previously unseen environments.

\begin{figure}[t]
    \centering
    \includegraphics[width=\linewidth]{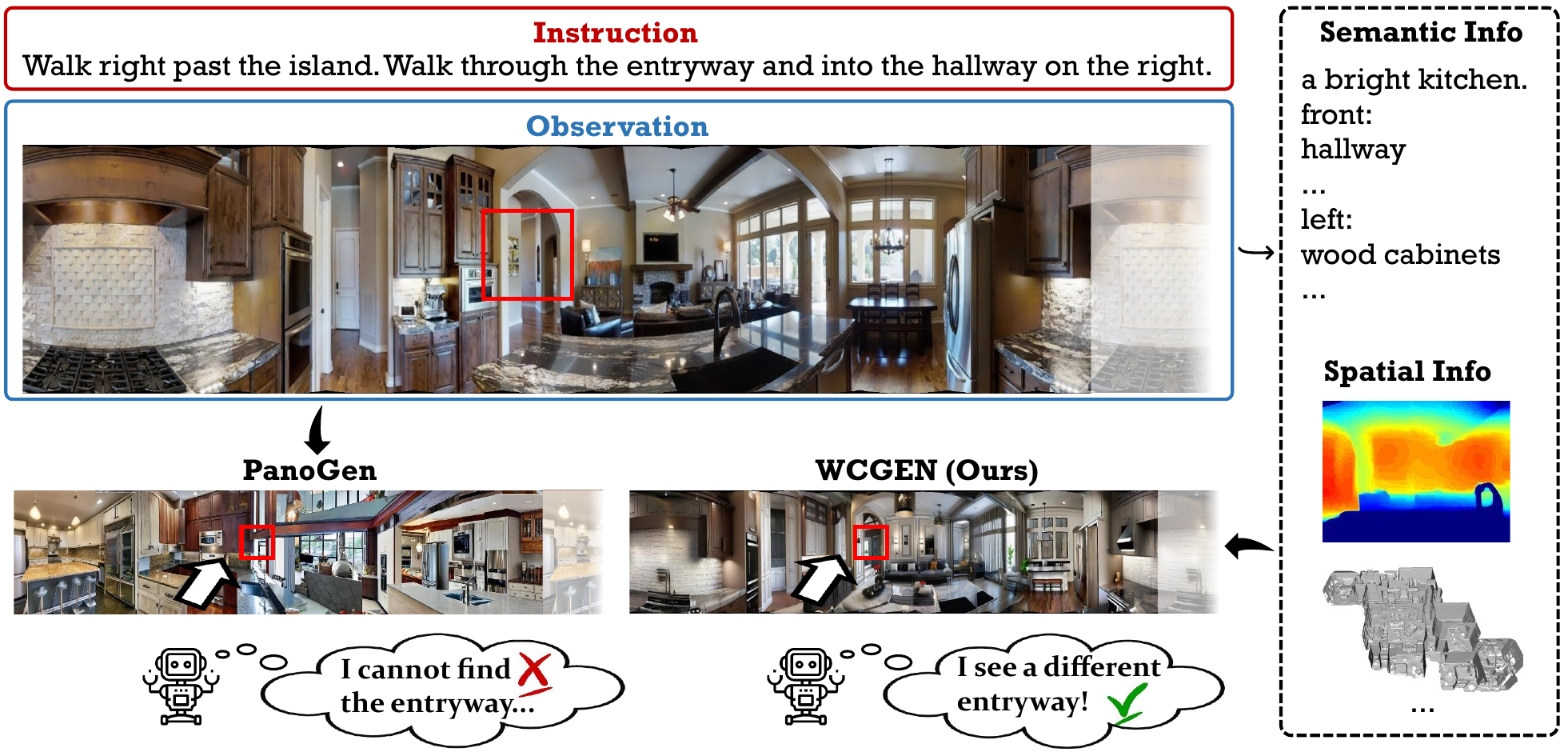}
    \caption{Comparison between our WCGEN and previous data augmentation.
    Benefiting from crucial semantic and spatial information, WCGEN generates world-consistent training environments for VLN agents, enabling them to better adapt and generalize to unseen scenarios.}
    \label{fig:teaser}
    \vspace{-20pt}
\end{figure}

A promising solution is to augment data for scaling up the diversity of training environments. 
Several approaches leverage new simulated 3D environments by systematically extracting visual observation and generating instructions through various sampling techniques \cite{26,27,47}. 
Some other methods instead focus on editing the Matterport3D environments by changing styles, object appearances, viewpoint locations, high-frequency textures, or other relevant features of visual observations  \cite{31,40,44}. 
However, all these studies exhibit insufficient diversity, constrained either by the volume of scenes available in new simulators or by the statistical pattern of Matterport3D.
Recently, PanoGen \cite{19} augments visual observations by leveraging the Stable Diffusion Model \cite{32} to recursively generate panoramic images from textual descriptions. Though having achieved impressive improvement in VLN tasks, PanoGen mainly pays attention to high-level semantics and ignores the essential knowledge of the 3D world, leading to severe world-inconsistency that could confuse the agent and impede navigation performance.
In conclusion, how to generate VLN data both diverse and world-consistent remains a challenge.

As depicted in Fig.~\ref{fig:teaser}, to address these shortcomings, we aim to achieve high-quality data generation by taking both diversity and world-consistency into account. 
Here \textit{world-consistency} delineates that all components in the synthesized panoramas share spatially consistent views coherently aligned with the real world. 
For VLN agents, world-consistency involves two aspects: trajectory level and viewpoint level.
At the trajectory level, world-consistency entails ensuring visual observations across various locations alongside a given trajectory are spatially coherent.
\ZY{At the viewpoint level, world consistency requires all perspective images to be seamlessly aligned and wraparound into a coherent panorama.}
We argue that world-consistency is paramount for synthesizing VLN datasets and serves a pivotal role in environmental perception and understanding.

To this end, we propose the world-consistent data generation (WCGEN), an efficacious data-augmentation framework satisfying both diversity and world-consistency, aiming at enhancing the generalization of agents to novel environments. 
The key insight of our approach is enriching diversity by employing the powerful image generation model while handling the world-consistency by incorporating camera projective geometry into the generation process.
Specifically, to achieve the above-mentioned two aspects of world-consistency, our method consists of two stages, the trajectory stage and the viewpoint stage, as \ZY{shown} in Fig.~\ref{fig:pipeline}.
First, in the trajectory stage, we leverage a point-cloud based technique to ensure the spatial coherency among viewpoints.
Then we adopt the diffusion model to generate the reference image that indicates the right navigation direction of each viewpoint along the trajectory.
Second, in the viewpoint stage, we adopt an accurate novel angle synthesis to guarantee the spatial coherency and wraparound consistency of the entire observation.
We subsequently outpaint the precedingly generated reference image into a complete panorama progressively.
During generation, we leverage a Large Language Model (LLM) to extract crucial information, such as layout descriptions and key objects, and feed them into the diffusion model as prompts.
Finally, we finetune a pre-trained visual-language model to generate new instructions for the trajectory. 
In conclusion, our methodology maintains world-consistency by accurately predicting viewpoint changes with 3D analysis during the generation procedure.
To validate the effectiveness of our proposed method, we measure it on the powerful VLN model DUET \cite{15} and carry out extensive experiments on two categories of VLN datasets: \textit{fine-grained navigation} (R2R, R4R \cite{42}) and \textit{coarse-grained navigation} (REVERIE \cite{21}, CVDN \cite{22}).
All experiments are conducted on the validation and test splits, illustrating the effectiveness of our method in enhancing generalization capability of VLN agents to unseen environments.
Empirical results show that our synthesized VLN data not only exhibits greater photorealism but also endows significant improvement on the VLN agents and outperforms state-of-the-art approaches in all navigation tasks.

\section{Related Works}

\subsection{Vision-and-Language Navigation}
Vision-and-Language Navigation (VLN) is a crucial but challenging embodied AI task requiring the agent to follow natural-language instructions to navigate through the visual environment \cite{1}. Early studies adopt LSTM as the backbone to better preserve historical observations \cite{2,3,4,5,6,7}. However, these methods suffer from severe information loss when capturing long-term dependencies as the navigation length increases. To solve this, researchers introduce the transformer-based architecture for the VLN task, utilizing its remarkable ability to extract multi-modal representations \cite{8,9,10,11,12,13,57,59}. 
Recently, many works delve into storing memories via topological graphs \cite{14,16,17,18,56}, 3D perception \cite{54,55} or knowledge \cite{58,60}, showing significant progress in performance. We choose the topology-based state-of-the-art method DUET as the baseline to explore our proposed data-augmentation algorithm.

\begin{figure*}[t]
    \centering
    \includegraphics[width=\linewidth]{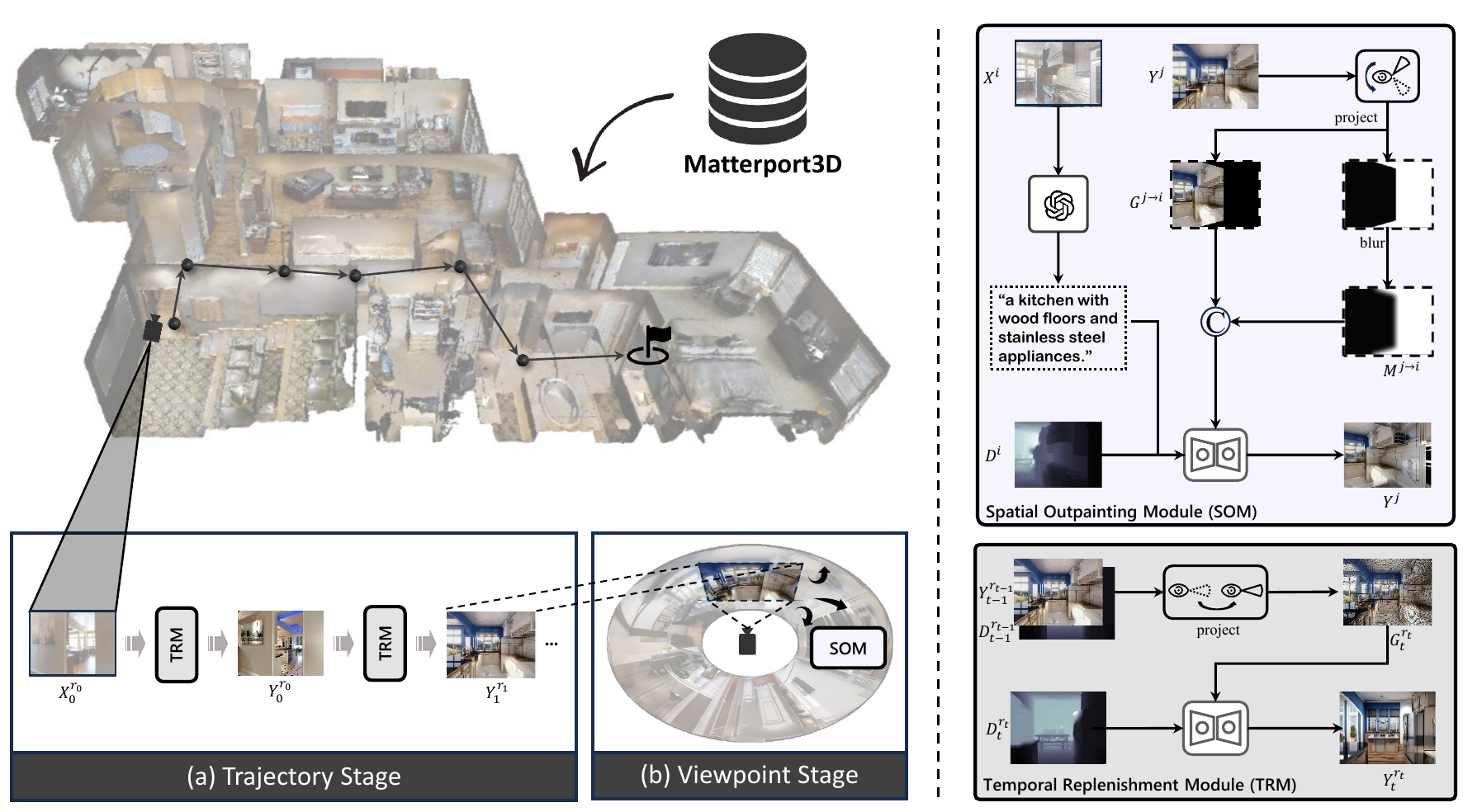}
    \caption{Framework of WCGEN. The input is a trajectory of viewpoints together with their panoramic observation, depth values, camera parameters, and positional information. Our strategy is to generate the VLN data in a two-stage manner. First, in the trajectory stage, we successively generate reference images along the navigation path while ensuring spatial coherency. Second, in the viewpoint stage, we iteratively outpaint the reference images to world-consistent panoramas.}
    \label{fig:pipeline}
\end{figure*}

\subsection{Data-augmentation for VLN}
One severe obstacle for VLN is the data scarcity issue due to the elaborate navigational data collection process.
Most existing datasets are constructed upon the Matterport3D dataset, which consists of only 61 environments for training.
To tackle this challenge, many works have been proposed to create more environments for VLN learning. 
Some studies aim to augment data based on the existing Matterport3D dataset by mixing up scenes or editing visual observations \cite{48}.
These methods are restricted by the Matterport3D environments and cannot provide significant enhancements on navigation.
The recently proposed PanoGen synthesizes new observations based on descriptive texts of original scenes. 
However, panoramic data generated by PanoGen faces the problem of 3D world inconsistency, which could provide ambiguous environmental information and lead to degrading navigation performance.
Some methods \cite{24,25}, which explore web resources to collect data for reducing training bias, are limited by the poor quality of web-sourced data, which is often fragmented and noisy, unlike structured VLN data.
Lastly, several approaches introduce other 3D environments to collect more training data. 
Despite having achieved impressive performance, they are formidable to further scale up considering that 3D environments are both expensive and scarce.
For a fair comparison, we do not include these methods in the main experiment.

\subsection{Diffusion Model-based Image Generation}

Denoising diffusion models ~\cite{song2019generative} have emerged as powerful generative models capable of capturing complex distributions in real-world 2D images. Recent works ~\cite{xue2024raphael} have showcased remarkable performance in image generation from text prompts. 
These models are widely utilized in various downstream applications, including image inpainting ~\cite{lugmayr2022repaint, corneanu2024latentpaint}, super-resolution ~\cite{wang2024exploiting}, video generation ~\cite{blattmann2023stable, wang2024videocomposer}, and even 3D asset generation ~\cite{poole2022dreamfusion, lin2023magic3d, wang2024prolificdreamer} due to their strong semantic priors learned from extensive training on large datasets.

Recently, latent diffusion models (LDM) and Stable Diffusion ~\cite{rombach2022high, podell2023sdxlimprovinglatentdiffusion} have been proposed that carry out the diffusion process in the latent space, greatly reducing computational complexity.
ControlNet ~\cite{zhang2023adding} introduces a new framework,
which extends the text-to-image (T2I) models to image-to-image translation tasks, enabling task-specific image generation. In this study, we leverage a pre-trained ControlNet model to synthesize novel view images guided by depth information.

\section{Methodology}
In this section, we present the world-consistent data generation (WCGEN) framework to create limitless world-consistent VLN data and benefit the generalization ability of agents.
Given a trajectory of viewpoints with corresponding panoramas and auxiliary information, our method creates both new visual observations and aligned instructions for the navigation training process.
To handle the world-consistency at both trajectory and viewpoint levels, we adopt a cascaded two-stage generation strategy, as illustrated in Fig.~\ref{fig:pipeline}.
First, in the trajectory stage, we progressively generate reference images along the navigation trajectory to ensure spatial coherency among viewpoints.
The reference image is defined as the perspective image that indicates the direction toward the next viewpoint along the given trajectory.
Second, in the viewpoint stage, we iteratively accomplish panoramic generation to better maintain the spatial coherency and wraparound consistency of the entire observation.
Finally, we gather the novel instructions aligned with the generated visual observations.

\subsection{The Trajectory Stage}
\label{ssec:trajectory}
To ensure the spatial coherency among viewpoints within the navigation trajectory, we propose the first stage, {\textit{i.e.} the trajectory stage, to synthesize the reference image of each viewpoint sequentially, as shown in Fig.~\ref{fig:pipeline}. 
To achieve this, we craft the Temporal Replenishment Module (TRM) which synthesizes the current reference image conditioned on previous reference images.

\textbf{Temporal Replenishment Module (TRM)}.
For the $t$-th viewpoint, supposing that generation of preceding reference image $Y_{t-1}^{r_{t-1}}$ has been accomplished, 
the proposed TRM takes $Y_{t-1}^{r_{t-1}}$ as the input to generate $Y_t^{r_t}$.
Here $Y_t^{r_t}$ denotes the $r_t$-th synthesized perspective image of the $t$-th viewpoint respectively.
We utilize $r_t$ to denote the perspective index of the reference image for the $t$-th viewpoint.
During the generation process, to ensure the spatial relationships between viewpoints, an accurate single-step view variation is required for a physically-grounded generation.
The point-cloud is capable of solving this issue by leveraging the 3D structural representation for capturing finer details and spatial dependencies.
Thus, we employ a point-cloud based projection technique to generate the guidance image $G_t^{r_t}$ which will guide the subsequent generation process of $Y_t^{r_t}$.
To support the point-cloud projection, we first calculate the depth information $\widetilde{D}_{t-1}^{r_{t-1}}$ of $Y_{t-1}^{r_{t-1}}$ with a pre-trained transformer-based depth estimator ~\cite{34}.
Based on camera parameters, we convert $Y_{t-1}^{r_{t-1}}$ and $\widetilde{D}_{t-1}^{r_{t-1}}$ into the guidance image $G_t^{r_t}$ with a point-cloud based method.
Specifically, given the 2D pixel coordinates $(u, v)$ in $Y_{t-1}^{r_{t-1}}$, we get the 3D point $P_w$:
\begin{equation}
\label{1}
    P = \mathbf{K}^{-1} \cdot \begin{bmatrix}
    u \cdot \widetilde{D}_{t-1}^{r_{t-1}}(u, v) \\
    v \cdot \widetilde{D}_{t-1}^{r_{t-1}}(u, v) \\
    \widetilde{D}_{t-1}^{r_{t-1}}(u, v)
    \end{bmatrix}
\end{equation}
\begin{equation}
    P_w = \mathbf{R} \cdot P + \mathbf{T}
\end{equation}
where $\mathbf{K}$ denotes the camera intrinsics, $\mathbf{R}$ denotes the rotation matrix, and $\mathbf{T}$ denotes the translation vector. 
Subsequently, we map $P_w$ into the new camera coordinate system of the next viewpoint:
\begin{equation}
    P' = \mathbf{R}_{t-1\rightarrow t}^{\mathsf{T}}\cdot(P_w - \mathbf{T}_{t-1\rightarrow t})
\end{equation}
where $\mathbf{T}_{t-1\rightarrow t}$ and $\mathbf{R}_{t-1\rightarrow t}$ denote the translation vector and rotation matrix from the $i-1$-th viewpoint to the $i$-th, respectively.
After that, we back-project $P'$ into the new coordinates of pixel domain $(u',v')$ of $G_t^{r_t}$, utilizing the inverse operation of Eqn. \eqref{1}.
In this manner, pixels from $Y_t^{r_t}$ are mapped to novel coordinates at the $t$-th viewpoint, resulting in the warped output $G_t^{r_t}$.
Considering that projected $G_t^{r_t}$ is fragmentary and involves obvious missing pixels, we need to convert it to a smooth and photorealistic image.
Direct inpainting on $G_t^{r_t}$ could yield ambiguous results since the average inter-viewpoint distance exceeds the maximum distance for ideal projection, causing dense grid-shaped missing regions distributed on $G_t^{r_t}$.
Instead, we adopt an image-to-image strategy that allows for drastic generation to overcome this obstacle.
We apply the ControlNet conditioned on $G_t^{r_t}$ and $D_t^{r_t}$ to execute the image-to-image generation and output $Y_t^{r_t}$.
Note that a reduced noise strength is utilized to constrain the extent of generation to prevent substantial degradation of the information stored in $G_t^{r_t}$.

Additionally, \ZY{it should be noted that} the paradigm is slightly different under the special circumstances below: 
1) $t=1$. 
We take the original perspective image $X_0^{r_0}$ as the input and omit the projection process, and 2) the perspective disparity between the $t$-th and $t-1$-th viewpoint is too large to leave any overlapping parts, leading to an invalid guidance image.
To handle this, we simply dump the guidance image and perform the text-to-image generation conditioned on the depth information since there are no in-consistent issues between the current and previous viewpoints.

\subsection{The Viewpoint Stage}
\label{ssec:viewpoint}
To guarantee layout coherence and multi-view consistency within the panoramic observation from a single viewpoint, we present the second stage, \textit{i.e.} the viewpoint stage, to outpaint the panorama for each viewpoint based on the reference image, as depicted in Fig.~\ref{fig:pipeline}.
Our paradigm is to outpaint each perspective image progressively, ensuring that newly generated images maintain spatial consistency with all previously generated images at each iteration.
For a specified viewpoint $t$, the viewpoint stage takes the reference image $Y_t^{r_t}$ and real images $X^{\{1,...,n\}/r}$ as inputs, where $n$ denotes the total number of perspective images.
Then we present the Spatial Outpainting Module to generate each perspective image $Y_t^i, i \in \{1,...,n\}/r$ step-by-step, in a traversal order.
For simplicity, we abbreviate $Y_t^i$ to $Y^i$ and omit the subscript. 
To better clarify the traversal order for generation, we introduce the target queue $Q$ which is a permutation of $\{1,...,n\}/r$.
We construct $Q$ as follows: Start with the reference image $Y^r$ and move rightward along the panorama, appending all perspective index within the same column including horizontal, upward, and downward into $Q$, until completing a loop and returning to $Y^r$.

\textbf{Spatial Outpainting Module (SOM)}
For each $i\in Q$, we first build the neighbor set $S^i$ by collecting all generated perspective images adjacent to the $i$-th perspective.
RM accepts the $S^i$, the real image $X^i$, and the corresponding depth information $D^i$ together as input and generates $Y^i$.
Given that there are substantial overlapping regions between different perspectives within a single viewpoint, it is imperative to reconstruct these regions with deliberation to better ensure the perspective changes adhere to established photogrammetric principles.
Our solution is to predict the appearance warp of common objects across various perspectives to approximate the effects of camera rotation.
To achieve this, we adopt a novel angle synthesis approach to \ZY{warp} $Y^j$ to the target view $i$, producing the guidance image $G^{j\rightarrow i}$ and the binary mask $M^{j\rightarrow i}$, both of which are instrumental in guiding the subsequent generation of $Y^i$.
Here we assume that there is only a single image $Y^j\in S^i$ for clarity.
Specifically, considering a pixel situated on $Y^j$ with homogeneous coordinates $\mathbf{v_\text{pixel}}$, we calculate the corresponding coordinates $\mathbf{v_\text{sphere}}$ on a projected unit sphere:
\begin{equation}
\mathbf{v}_{\text{sphere}} = \frac{\mathbf{K}^{-1}\cdot\mathbf{v_\text{pixel}}}{\|\mathbf{K}^{-1}\cdot\mathbf{v_\text{pixel}}\|}
\end{equation}
where $\mathbf{K}$ denotes the camera intrinsics. 
Next, we conduct the perspective rotation:
\begin{equation}
\mathbf{v}_{\text{sphere}}' = \mathbf{R}_{j\rightarrow i}\cdot\mathbf{R}\cdot\mathbf{v}_{\text{sphere}}
\end{equation}
where $\mathbf{R}_{j\rightarrow i}$ represents the rotation matrix that transforms from the perspective $j$ to $i$, while $\mathbf{R}$ denotes the extrinsic rotation matrix.
Finally, we conduct the rasterization to turn $\mathbf{v}_{\text{sphere}}'$ back into pixel coordinates $\mathbf{v}_{\text{pixel}}'$ located on the guidance image $G^{j\rightarrow i}$. 
Meanwhile, $M^{j\rightarrow i}$ is constructed by applying a binary thresholding operation to $G^{j\rightarrow i}$, setting all pixels with non-zero values in $G^{j \rightarrow i}$ to one in $M^{j \rightarrow i}$ and others to zero.
To soften the transition regions during generation, we blur the edges of $M^{j\rightarrow i}$ and acquire the blurred mask $\hat{M}^{j\rightarrow i}$.
In the end, we leverage the ControlNet to conduct image outpainting based on the $G^{j\rightarrow i}$, along with depth information $D^i$, $\hat{M}^{j\rightarrow i}$ to generate the output $Y^i$.
Here the textual prompts are acquired from GPT-4 \cite{62} to gather the crucial information for $X^i$.
We prompt the LLM with two questions to extract the coarse-grained layout description and fine-grained object-level information.
The exact prompts are provided in the supplementary.
When multiple images are involved in $S_i$, these neighboring images originate from various directions relative to $X^i$, resulting in a minimal intersection in their overlapping areas with $X^i$.
Therefore, the previously described outpainting paradigm can be extended to multiple images by performing a weighted sum of the guidance images and their corresponding masks. 
More details on our method are presented in the supplementary.
\subsection{Instruction Generation}
\label{ssec:instruction}
To achieve a better cross-modal alignment, it is essential for us to synthesize new instructions for the previously generated observations. 
Inspired by previous works\cite{19,31}, we observe that instruction texts of VLN tasks can be regarded as describing the scenes along a certain navigation path to some extent, which shares a very similar pattern with video captioning.
Therefore, we finetune the multimodal mPLUG-2 \cite{50}, the state-of-the-art video captioning approach, with the Matterport3D training set to generate novel instructions. 
By doing so, we acquire the instructions matched with the generated panoramas, which can be beneficial for improving the vision-text grounding capability of VLN agents and thereby enhancing their generalization ability.

\begin{table*}\large
  \centering \resizebox{2.0\columnwidth}{!}{
  \centering
  \begin{tabular}{c|cccc|cccc|cccc}
    \toprule
   \multirow{3}*{\textbf{Methods}} & \multicolumn{8}{c|}{\textbf{Room-to-Room}} & \multicolumn{4}{c}{\textbf{Room-for-Room}} \\  
   \cmidrule{2-13}
   & \multicolumn{4}{c|}{\textbf{Validation Unseen}}  & \multicolumn{4}{c|}{\textbf{Test}} &   \multicolumn{4}{c}{\textbf{Validation Unseen}} \\ 
   \cmidrule{2-13}
    & TL & NE $\downarrow$ & SR $\uparrow$ & SPL $\uparrow$ & TL & NE $\downarrow$ & SR $\uparrow$ & SPL $\uparrow$ & TL & NE $\downarrow$ & SR $\uparrow$ & SPL $\uparrow$ \\
    \midrule
    \midrule
    Human & - & - & - & - & 11.85 & 1.61 & 86.0 & 76.0 & - & - & - & - \\
    \midrule
    EnvDrop{$\dag$}~\cite{37} & 10.70 & 5.22 & 52.0 & 48.0 & 11.66 & 5.23 & 51.0 & 47.0 & - & 9.18 & 34.7 & 21.0 \\
    PREVALENT~\cite{38} & 10.19 & 4.71 & 58.0 & 53.0 & 10.51 & 5.30 & 54.0 & 51.0 & - & - & - & - \\
    VLN$\protect\circlearrowright$BERT~\cite{10} & 12.01 & 3.93 & 63.0 & 57.0 & 12.35 & 4.09 & 63.0 & 57.0 & - & 6.48 & 42.5 & 32.4 \\
    HAMT~\cite{8} & 11.46 & 3.62 & 66.0 & 61.0 & 12.27 & 3.93 & 65.0 & 60.0 & - & 6.09 & 44.6 & - \\ 
    REM{$\dag$}~\cite{30} & 12.44 & 3.89 & 63.6 & 57.9 & 13.11 & 3.87 & 65.2 & 59.1 & - & 6.21 & 46.0 & 38.1 \\ 
    AirBert{$\dag$}~\cite{45} & 11.78 & 4.10 & 62.0 & 56.3 & 12.41 & 4.13 & 61.5 & 57.1 & - & - & - & - \\
    EnvEdit{$\dag$}~\cite{31} &  12.13 & 3.22 & 67.9 & 62.9 & - & - & - & - & - & - & - & - \\
    DUET~\cite{15} & 13.94 & 3.31 & 72.0 & 60.0 & 14.73 & 3.65 & 69.0 & 59.0 & 21.04 & 6.06 & 46.6 & 41.9 \\
    MARVAL{$\dag$}~\cite{44} & 10.15 & 4.06 & 65.2 & 61.0 & 10.22 & 4.18 & 62.1 & 58.2 & - & - & - & - \\ 
    SE3DS{$\dag$}~\cite{40} & - & 3.29 & 69.0 & 62.0 & - & 3.67 & 66.0 & 60.0 & - & - & - & - \\
    Lily{$\dag$}~\cite{24} & 14.58 & \underline{2.90} & 74.0 & 62.3 & 16.13 & 3.44 & \underline{72.1} & 60.7 & - & - & - & - \\
    FDA{$\dag$}~\cite{44} & 13.68 & 3.02 & 72.1 & 64.0 & 14.76 & 3.41 & 68.7 & 62.4 & - & - & - & - \\
    PanoGen{$\dag$}~\cite{19} & 13.40 & 3.03 & \underline{74.2} & \underline{64.3} & 14.38 & \underline{3.31} & 71.7 & \underline{63.9} & 18.62 & \underline{6.02} & \underline{47.8} & \underline{44.3} \\
    \midrule
    WCGEN{$\dag$}+DUET & 12.57 & \textbf{2.89} & \textbf{75.0} & \textbf{65.8} & 13.00 & \textbf{3.22} & \textbf{72.7} & \textbf{64.5} & 18.87 & \textbf{5.89} & \textbf{48.5} & \textbf{45.3} \\
    \bottomrule
  \end{tabular}
  }
  \caption{Coarse-grained navigation results on R2R and R4R datasets. The best results are in bold while the second are underlined. Approaches with the notation $\dag$ indicate that they use the data augmentation.}
  \label{Tab_1}
\end{table*}

\begin{table}[t]
  \centering
  \resizebox{1.0\columnwidth}{!}{
  \begin{tabular}{c|c|c|cccc}
  \toprule
  \multirow{3}*{\textbf{Methods}} & \multicolumn{2}{c|}{\textbf{CVDN}} & \multicolumn{4}{c}{\textbf{REVERIE}} \\  
   \cmidrule{2-7}
   & \textbf{Validation Unseen} & \textbf{Test} & \multicolumn{4}{c}{\textbf{Validation Unseen}} \\ 
   \cmidrule{2-7}
    & GP $\uparrow$ & GP $\uparrow$ & OSR $\uparrow$ & SPL $\uparrow$ & RGS $\uparrow$ & RGSPL $\uparrow$ \\
    \midrule\midrule
    PREVALENT~\cite{38} & 3.15 & 2.44 & - & - & - & - \\
    VLN$\protect\circlearrowright$BERT~\cite{10} & - & - & 35.02 & 24.90 & 24.90 & 18.77 \\
    HAMT~\cite{8} & 5.13 & 5.58 & 36.84 & 30.20 & 18.92 & 17.28 \\
    DUET~\cite{15} & 5.50 & - & 51.07 & \underline{34.14} & 31.70 & \textbf{22.89} \\
    PanoGen~\cite{19} & \underline{5.93} & \underline{7.17} & \underline{51.19} & 33.44 & \textbf{32.80} & \underline{22.45} \\
    WCGEN+DUET & \textbf{6.47} & \textbf{7.18} & \textbf{52.23} & \textbf{34.33} & \underline{31.82} & 22.33 \\
    \bottomrule
  \end{tabular}
  }
  \caption{Fine-grained navigation results on CVDN and REVERIE.}
  \label{Tab_2}
\end{table}

\begin{figure*}[t]
    \centering
    \centering
    \includegraphics[width=\linewidth]{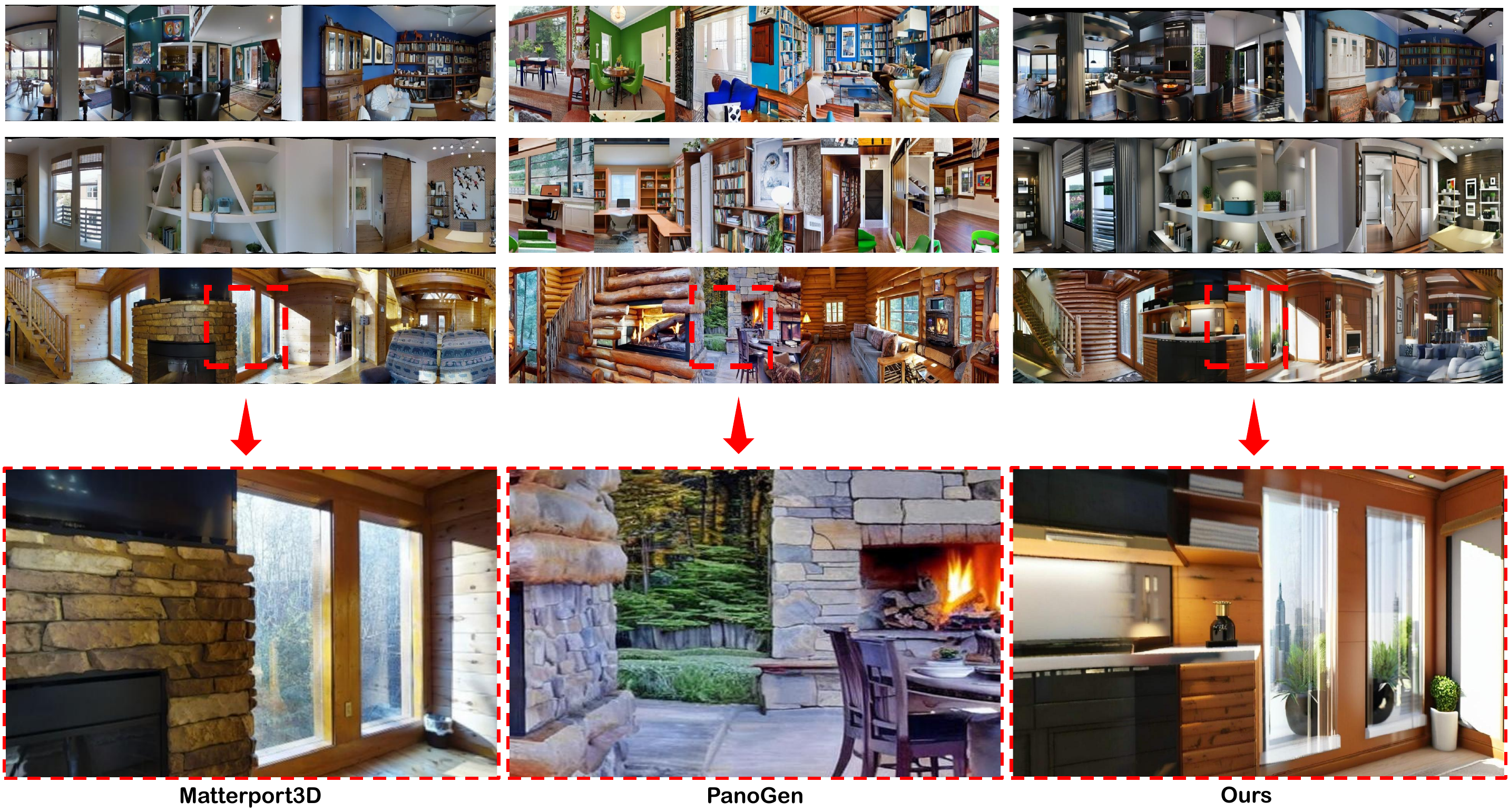}
    \caption{Qualitative results for panoramas and perspective images. We visualize three viewpoints and observe that our results exhibit more reasonable layout and consistent semantics than PanoGen.}
    \label{Fig_qualitative}
\end{figure*}

\section{Experiments}
%
%
%

\subsection{Datasets}
We evaluate our approach on two categories of VLN datasets: \textit{fine-grained navigation} (R2R \cite{20} and R4R \cite{42}) and \textit{coarse-grained navigation} (REVERIE \cite{21} and CVDN \cite{22}). 
These datasets are all created based on the Matterport3D \cite{23}, which includes 90 photo-realistic environments with 10,567 egocentric panoramas in total.
\begin{itemize}
    \item \textbf{R2R} is the first proposed VLN benchmark, composed of 7,189 direct-to-goal shortest paths, each associated with 3 human-annotated navigation instructions. It is split into training, test, validation seen, and validation unseen with 61, 18, 56, and 11 environments. Environments of the test and validation unseen sets do not appear in the training set.
    \item \textbf{R4R} augments the R2R by joining two existing paths end-to-end if the distance between the ending viewpoint of the first path and the starting viewpoint of the other path is within a certain threshold. With longer and more divergent paths, R4R affords a more comprehensive assessment of the capacity of an agent to follow instructions.
    \item \textbf{REVERIE} introduces a more complicated and varied task, requiring agents to navigate and identify a remote object with high-level descriptions. Such descriptions are mostly ambiguous and only provide the target location and object.
    \item \textbf{CVDN} takes a step further and demands agents to infer and explore where the target object is located based on a dialog history.
\end{itemize}

\begin{table}[t]
\centering
\tiny
\resizebox{0.73\columnwidth}{!}
{
    \begin{tabular}{c|cc}
    \toprule
    \textbf{Methods} & \textbf{IS$\uparrow$} & \textbf{FID$\downarrow$} \\
    \midrule\midrule
    Stable Diffusion & 1.58 & 29.79\\
    EnvEdit & 1.41 & 30.24\\
    PanoGen & 1.62 & 29.65\\
    WCGEN & \textbf{2.23} & \textbf{27.54}\\
    \bottomrule
    \end{tabular}
}
\caption{Quantitative results for panoramic quality evaluation.}
\label{Tab_quantitative}
\end{table}

\subsection{Metrics}
We utilize standard metrics~\cite{43} to measure navigation performance as follows: 
(1) Trajectory Length (\textbf{TL}): the navigation path length in meters on average;
(2) Navigation Error (\textbf{NE}): the distance between the agent’s final position and the target in meters on average;
(3) Success Rate (\textbf{SR}): the proportion of trajectories that successfully reach the destination with NE less than 3 meters;
(4) Success weighted by Path Length (\textbf{SPL}): the success rate normalized by the ratio between the length of the shortest path and the predicted path, balancing both SR and TL;
(5) Goal Progress (\textbf{GP}): the total distance that agents navigate toward the destination;
(6) Oracle Success Rate (\textbf{OSR}): the success rate when oracle information is accessible.
Among these, SPL is widely regarded as the primary measure of navigation performance~\cite{43}.
Moreover, we also adopt the following metrics to evaluate the object grounding task:
Remote Grounding Success (\textbf{RGS}): the proportion of tasks that successfully locate the target object (IoU between the predicted bounding box and the ground truth is larger than $0.5$);
and Remote Grounding Success weighted by Path Length (\textbf{RGSPL}): RGS normalized by the path length.

\subsection{Implementation}
Our method is implemented with PyTorch and the official release of CLIP ViT-B/16~\cite{53}, ControlNet-depth\cite{36}, Stbale Diffusion XL~\cite{52}, GPT-4, and mPLUG-2~\cite{50}.
Here GPT-4 is adopted through OpenAI’s official API.
CLIP ViT-B/16 is applied to extract the visual features of perspective images similar to ~\cite{19}.
Besides, our approach is evaluated on the representative method DUET. We follow the same training schemes as the official code, including pre-training on several proxy tasks and fine-tuning on VLN datasets. 
See the supplement for more details.


\subsection{Experimental Results}
\subsubsection{Fine-grained navigation}
As presented in Table~\ref{Tab_1}, we compare our proposed WCGEN with other VLN methods on the R2R and R4R datasets.
By training DUET with our augmented data, we yield improvements of $5.8\%$, $5.5\%$ and $3.4\%$ on the R2R validation unseen, R2R test and R4R validation unseen splits respectively in terms of SPL.
%
These large increments in performance demonstrate that learning from our environments significantly enhances the generalization of VLN agents to unseen environments. 
Moreover, our approach also outperforms the state-of-the-art data augmentation method PanoGen by $1.5\%$, $0.6\%$, and $1.0\%$ in SPL on the three splits, respectively. 
In conclusion, utilizing our augmented world-consistent environments significantly enhances the generalization of VLN agents to unseen environments.

\subsubsection{Coarse-grained navigation}
Table~\ref{Tab_2} shows experimental results on the CVDN and REVERIE datasets. 
As for CVDN, our WCGEN outperforms PanoGen by $9.1\%$ in GP on the validation unseen split. 
This relative elevation is much larger than R2R and R4R, demonstrating that our method is especially effective for complex situations with under-specified and long instructions.
As for REVERIE dataset with high-level instructions, our approach exhibits superior performance over other methods in all navigation metrics, including SPL where the PanoGen performs worse than the baseline method. 
This indicates that agents trained with our augmented data can better handle navigation with ambiguous instructions.
Nevertheless, our method demonstrates inferior performance compared to DUET on  RGSPL. Through systematic analysis of failure cases where DUET succeeds but our approach falters, we identify a common characteristic: these instances predominantly involve small-scale target objects (e.g., candles). We attribute this phenomenon primarily to RGSPL's evaluation protocol emphasizing object detection precision, while our framework prioritizes optimizing fundamental navigation skills over detailed object generation.

\subsubsection{Generation quality evaluation}
Fig.~\ref{Fig_qualitative} presents qualitative comparisons regarding image quality between our method and baseline approaches. The visual comparisons demonstrate that WCGEN-generated panoramas exhibit better spatial coherence and more natural visual-distortion patterns compared to PanoGen. Besides, our framework effectively maintains geometric consistency through seamless $360$\degree integration. Notably, an enlarged region of the red-dotted box in Fig.~\ref{Fig_qualitative} Row 3 provides a detailed demonstration of our structural preservation capabilities. In the scene with the caption "\textit{a living room with a fireplace}", our method preserves the essential layouts while modifying both semantic contents (e.g., adding a shelf with towels and bottles) and visual appearance (e.g., the color of the fireplace changes from light earthy to yellow). Modifications in both the semantic and appearance levels can ensure world consistency and expand the diversity of the generated data.
Moreover, we conduct evaluations of the generated perspective images using both Inception Score (\textbf{IS}) and Fréchet Inception Distance (\textbf{FID}), with detailed comparison results shown in Table~\ref{Tab_quantitative}. Notably, our framework achieves superior performance compared to existing VLN data generation methods across both evaluation metrics. These results collectively demonstrate that our world-consistent generation framework produces more photorealistic and physically-grounded results, thereby enhancing VLN agents' ability to comprehend and interact with the real world.

\begin{table}[t]
  \centering
  \resizebox{1.0\linewidth}{!}
  {
    \begin{tabular}{cc|cc|ccc|cccc}
    \toprule
    \multicolumn{4}{c|}{\textbf{Trajectory Stage}} & \multicolumn{3}{c|}{\textbf{Viewpoint Stage}} & \multicolumn{4}{c}{\textbf{Validation Unseen}} \\
    \midrule
    IM & IM$_{w/o}$ & FM & FM$_{w/o}$ & RM & RM$_{-d}$ & RM$_{w/o}$ & TL & NE~$\downarrow$ & SR~$\uparrow$ & SPL~$\uparrow$ \\
    \midrule\midrule
    \usym{2713} & & \usym{2713} & & \usym{2713} & & & 12.57 & \textbf{2.89} & \textbf{75.0} & \textbf{65.8} \\
    \midrule
    & \usym{2713} & \usym{2713} & & \usym{2713} & & & 13.58 & 2.91 & 74.8 & 65.3 \\
    \usym{2713} & & & \usym{2713} & \usym{2713} & & & 13.64 & 2.95 & 73.3 & 64.9 \\
    & \usym{2713} & & \usym{2713} & \usym{2713} & & & 14.68 & 3.01 & 73.5 & 64.1 \\
    \midrule
    \usym{2713} & & \usym{2713} & & & \usym{2713} & & 12.64 & 3.07 & 71.6 & 62.8 \\
    \usym{2713} & & \usym{2713} & & & & \usym{2713} & 13.07 & 3.19 & 72.2 & 63.1 \\
    \bottomrule
    \end{tabular}%
  }
  \caption{Ablation study on the effectiveness of modules. IM$_{w/o}$ means dropping IM and using real data as the first reference image; FM$_{w/o}$ means dropping FM and generating reference images independently; $R_{-d}$ stands for outpainting without depth information in RM; $R_{w/o}$ stands for dropping RM and generating perspective images independently.}
  \label{Tab_3}
\end{table}

\begin{table}[t]
\centering
\resizebox{1.0\linewidth}{!}{
\begin{tabular}{lcccc}
\toprule
\textbf{Methods} & TL & NE $\downarrow$ & SR $\uparrow$ & SPL $\uparrow$ \\
\midrule
DUET~\cite{15} & 13.94 & 3.31 & 72.0 & 60.0 \\
WCGEN & 12.57 & \textbf{2.89} & \textbf{75.0} & \textbf{65.8} \\
WCGEN (original) & 14.07 & 3.18 & 73.7 & 62.5 \\
\bottomrule
\end{tabular}
}
\caption{Ablation study on instruction generation. WCGEN (original) means using original instructions from R2R during training.}
  \label{tab:exp_instructions}
\end{table}

\subsection{Ablation Study}
In this section, we evaluate the effectiveness of WCGEN's component and analyze the influence of hyperparameters. All experiments are conducted on R2R validation unseen.

\subsubsection{Panoramic Generation}
We conduct an ablation analysis of key components in our two-stage generation framework, as shown in Table~\ref{Tab_3}. Row 1 presents the full configuration, while subsequent rows demonstrate performance degradation when selectively removing individual modules. The consistent performance decline across all ablated configurations confirms that each module serves as a critical functional component of WCGEN. Our experiments quantify two essential elements: (1) depth-aware rendering contributes to spatial coherence, and (2) the multi-view angle synthesis mechanism ensures geometric continuity. The removal of either component induces statistically significant performance deterioration, with minimum reductions of 4.1\% and 3.8\% in SPL and SR respectively.

\subsubsection{Instruction Generation}
As shown in Table~\ref{tab:exp_instructions}, we demonstrate the effectiveness of instruction generation.
The results indicate that using refined instructions can yield a 5.2\% improvement in SPL compared with original instructions.
This enhancement is primarily attributed to the consistency of the reference images ensured during the trajectory stage, which are taken as input to the mPLUG-2 model for instruction generation.

\begin{figure*}[t]
    \centering
    \includegraphics[width=\linewidth]{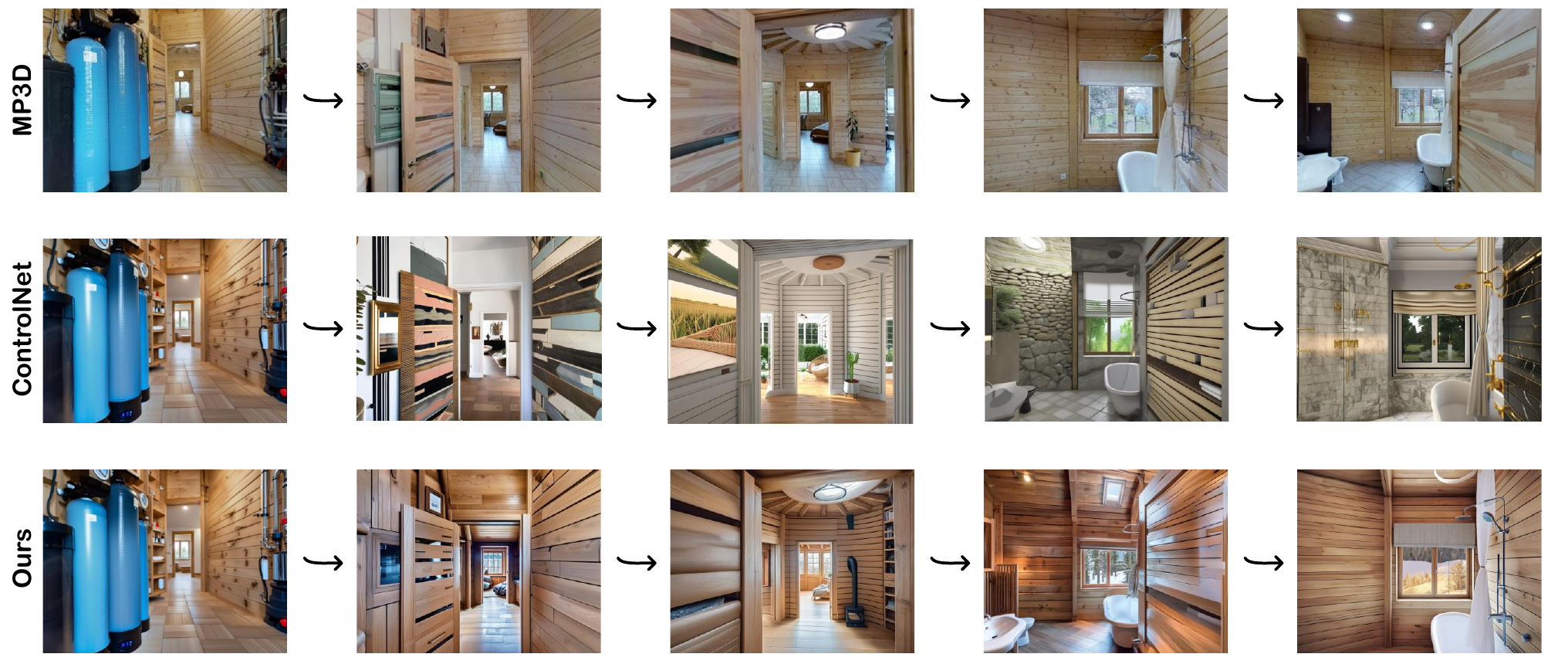}
    \caption{\textbf{Visualization of temporal consistency }. Our trajectory-based generation (the third line) exhibits more temporally consistency in styles and semantics than independently generated images (the second line).}
    \label{fig:trajectory}
    \vspace{-15pt}
\end{figure*}

\begin{figure*}[t]
    \centering
    \centering
    \includegraphics[width=\linewidth]{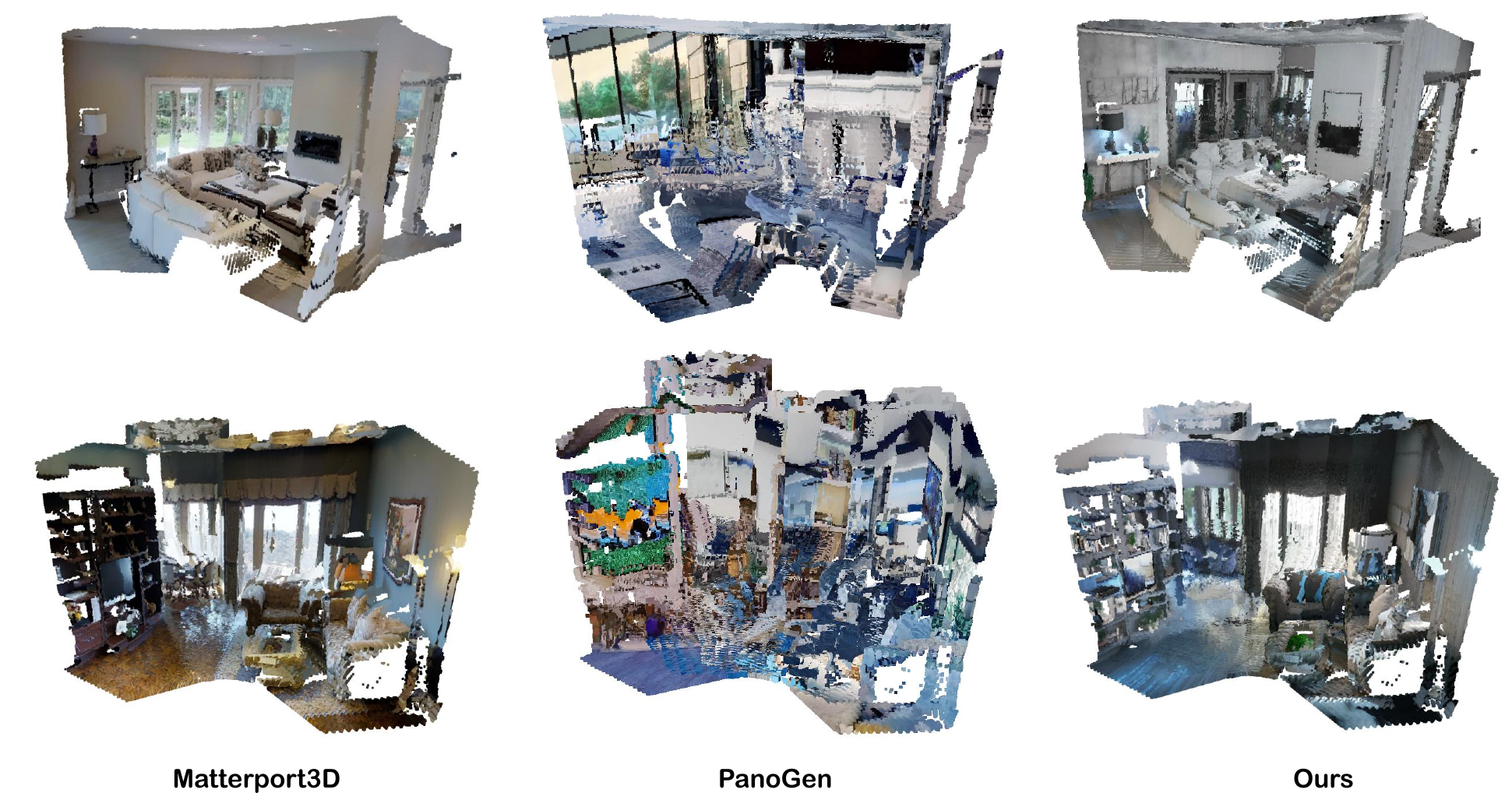}
    \caption{\textbf{Visualization of spatial consistency}. We perform depth estimation on the generated RGB images and further establish the 3D point clouds. Our generation is capable of reserving realistic layouts and exhibiting clear geometry of indoor scenes.}
    \label{Fig_5}
    \label{fig:pointcloud}
    \vspace{-10pt}
\end{figure*}

\begin{table}[t]
  \centering
  \resizebox{1.0\linewidth}{!}
  {
    \begin{tabular}{c|c|cccc}
    \toprule
    \textbf{pre-train} & \textbf{finetune} & TL & NE~$\downarrow$ & SR~$\uparrow$ & SPL~$\uparrow$ \\
    \midrule
    \usym{2713} & \usym{2713} & 12.57 & \textbf{2.89} & \textbf{75.0} & \textbf{65.8} \\
    \usym{2713} & \usym{2717} & 14.56 & 3.06 & 72.9 & 61.9 \\
    \usym{2717} & \usym{2713} & 13.71 & 2.93 & 74.0 & 63.5 \\
    \usym{2717} & \usym{2717} & 13.94 & 3.31 & 72.0 & 60.0 \\
    \bottomrule
    \end{tabular}
    }
  \caption{Ablation study on training phrases.}
  \label{tab:exp_other_phases}
\end{table}

\begin{table}[t]
\centering
\resizebox{1.0\columnwidth}{!}{
\begin{tabular}{c|ccccc}
    \toprule
    \textbf{Trajectory Number} & TL & NE $\downarrow$ & SR $\uparrow$ & SPL $\uparrow$ \\
    \midrule
    1000 & 13.96 & 2.99 & 74.4 & 64.1   \\
    2000 & 13.48 & 3.15 & 73.0 & 63.8 \\ 
    3000 & 12.92 & 2.95 & 73.9 & 63.9  \\
    4676 & 13.37 & 2.99 & 74.5 & 64.2 \\ 
    5000 & 12.57 & \textbf{2.89} & \textbf{75.0} & \textbf{65.8} \\
    \bottomrule
  \end{tabular}
  }
\caption{Ablation study on trajectory number.}
\label{tab:trajectory_number}
\end{table}

\subsubsection{Training Scheme}
In this subsection, we delve into the impact of different settings in the training scheme of VLN agents.
Firstly, we present the results validating the importance of our augmented data at different training phases in Table~\ref{tab:exp_other_phases}. Our results reveal that incorporating augmented data across all phases yields significant improvements over partial usage. Specifically, omitting augmented data during the finetuning phase causes higher performance degradation in SPL compared to its exclusion in the pre-training phase (5.9\% vs 3.5\%).
This is mainly because proxy tasks involved in the pre-training phase lack a strong connection with VLN tasks while our augmented data is specifically designed and organized for navigation purposes. Consequently, the fine-tuning phase benefits more from our augmentation.
Secondly, we explore the impact of data scales by randomly fetching a given number of generated trajectories for training.
As shown in Table~\ref{tab:trajectory_number}, the SR and SPL performance demonstrate a sustained increase as the scale of trajectories expands since the number exceeds $1000$.
Note that the increase does not plateau even when the number of trajectories exceeds 4,676, which is the total number of trajectories in the R2R dataset. 
This observation highlights the substantial potential of our approach, suggesting that additional advancement could be realized with further scaling.

\subsection{Visualization Analysis}
In this subsection, we provide more visualization results to demonstrate the superiority of our approach.
First, to illustrate the temporal consistency and effectiveness of the trajectory stage, we compare our synthesized reference images with the results of generating each viewpoint independently.
As depicted in Fig.~\ref{fig:trajectory}, our generation is capable of ensuring the consistency of both appearance and style along the trajectory much better (for instance, all bright yellow wooden interiors). 
Considering that agents are required to follow a single instruction in one episode, consistent scenes are beneficial for them to learn to observe and memorize the surroundings when making actions.
Second, we also verify the spatial coherency of our generation by visualizing 3D structures.
Specifically, we utilze the Metric3D v2 model \cite{61} to calculate the depth map of perspective images and convert the RGB-D to 3D point clouds.
After that, we merge these 3D point clouds from different views through projection.
As shown in Fig.~\ref{fig:pointcloud}, the visualized point clouds are capable of reserving the basic spatial structures of rooms and exhibiting clear geometry.



\section{Conclusion}
In this paper, we present the world-consistent data generation (WCGEN), a VLN data-augmentation framework aiming at enhancing the generalizations of agents to novel environments. 
To achieve this, we design a two-stage generation paradigm.
In the first trajectory stage, we apply the Temporal Replenishment Module (TRM) to sequentially generate reference images along the navigation trajectory to ensure spatial coherency among viewpoints. 
In the second viewpoint stage, we iteratively accomplish panoramic generation with the Spatial Outpainting Module (SOM) to better maintain the spatial coherency and wraparound consistency of the entire observation. 
Finally, we finetune a visual-language model to regenerate instructions.
By training with augmented data, we achieve new state-of-the-art results on various benchmarks with different categories of instructions.
This demonstrates that WCGEN is beneficial for enhancing the generalizing ability of VLN agents to previously unseen environments.


\bibliographystyle{IEEEtran}
\bibliography{egbib}


 




\vfill

\end{document}